\documentclass[conference]{IEEEtran}
\usepackage[utf8]{inputenc}

\usepackage{comment}
\usepackage{algorithm}
\usepackage{algorithmic}
\usepackage{cite}
\usepackage{amsmath,amssymb,amsfonts}
\usepackage{algorithmic}
\usepackage{graphicx}
\usepackage{textcomp}
\usepackage{xcolor}
\def\BibTeX{{\rm B\kern-.05em{\sc i\kern-.025em b}\kern-.08em
    T\kern-.1667em\lower.7ex\hbox{E}\kern-.125emX}}

\usepackage{graphicx}
\usepackage{subcaption}

\begin{document}

\title{ParaFIS:A new online fuzzy inference system based on parallel drift anticipation
{\footnotesize \textsuperscript{}}
\thanks{}
}

\author{\IEEEauthorblockN{Clement Leroy}
\IEEEauthorblockA{\textit{Intuidoc Team} \\
\textit{Univ Rennes, CNRS, IRISA}\\
Rennes, France \\
clement.leroy@irisa.fr}
\and
\IEEEauthorblockN{Eric Anquetil}
\IEEEauthorblockA{\textit{Intuidoc Team} \\
\textit{Univ Rennes, CNRS, IRISA}\\
Rennes, France \\
eric.anquetil@irisa.fr}
\and
\IEEEauthorblockN{Nathalie Girard}
\IEEEauthorblockA{\textit{Intuidoc Team} \\
\textit{Univ Rennes, CNRS, IRISA}\\
Rennes, France \\
nathalie.girard@irisa.fr}
}

\maketitle

\begin{abstract}
This paper proposes a new architecture of incremental fuzzy inference system (also called Evolving Fuzzy System - EFS). In the context of classifying data stream in non stationary environment, concept drifts problems must be addressed. Several studies have shown that EFS can deal with such environment thanks to their high structural flexibility. These EFS perform well with smooth drift (or incremental drift). The new architecture we propose is focused on improving the processing of brutal changes in the data distribution (often called brutal concept drift). More precisely, a generalized EFS is paired with a module of anticipation to improve the adaptation of new rules after a brutal drift.  The proposed architecture is evaluated on three datasets from UCI repository where artificial brutal drifts have been applied. A fit model is also proposed to get a "reactivity time" needed to converge to the steady-state and the score at end. Both characteristics are compared between the same system with and without anticipation and with a similar EFS from state-of-the-art. The experiments demonstrates improvements in both cases.
\end{abstract}
\begin{IEEEkeywords}
Evolving Fuzzy System EFS, Non Stationary Environment NSE,brutal drift adaptation, long-life learning.
\end{IEEEkeywords}

\section{Introduction}

More and more data are generated by stream over a long period leading to the necessity for classification methods to include long-life learning in their algorithm to stay reliable in time. 
As an example, a command-gesture system \cite{Bouillon2017Nov}, where the user draws a gesture to apply a command, produces this kind of data stream. The set of gestures is chosen by the user and can change at any time. 
For example a novice user often draw slowly a gesture, and once he is used to make it, he can speed up the gesture leading to a shift of the target concept in the feature space. This modification is called \emph{concept drift}. 
Such environment is said to be non stationary. It means that the probability distribution that generates data (here the user) depends on time. To maintain high performance in gestures recognition over time, the classification system must adapt its parameters and/or structure to the concept drift. Thus, the learning process has to be \emph{incremental}. For each new data, the system extends its knowledge (\textit{i.e.}, it learns and adapts its parameters). Moreover, to get a constant complexity in time and avoid a memory overflow, along a long-life learning, data should not be saved.
\\
Recent researches have shown that Evolving Fuzzy Systems - EFS (also called Incremental Fuzzy Inference System) could deal with such environment thanks to their high structural flexibility \cite{Lughofer2018,angelov2011}. The structure (number of rules, and antecedents/consequent parameters) of these fuzzy rule-based systems evolve with the stream of data.
Especially, we distinguish between two scales of adaptation to concept drift. One concerns the adaptation of the rules' parameters and tackles incremental drifts (or smooth drifts). The other relates to the adaptation of the structure, by the addition or deletion of rules, and tackles brutal drifts (brutal shifts of the data distribution).  
Both adaptions use two main algorithmic approaches. The first one relies on temporal (adaptive) sliding windows where only the most recent data are taking into account.
As an example, the fuzzy windows concept drift adaptation method (FW-DA) \cite{Liu2017Jul} relies on the method and shows promising results outperforming state-of-the-art.
The second one weights the data according to their age and interest. An example is \cite{Manuel2013} which uses a decremental learning on the premise and conclusion parts with the Differed Directional Forgetting (DDF). 
However, both approaches have to figure out a "forgetting" parameter which controls the relevance of the old data. If the forgetting parameter is too high, the system will be less stable leading to lower performance. If the parameter is too low, the system will not be reactive to the change in the environment leading to bad performance. This issue is often called \emph{the plasticity-stability dilemma}.
This dilemma can result in instability, particularly with EFS where the number of rules, their size and their reactivity greatly depend on parameters given \textit{a priori}.
To prevent instability, \cite{Lughofer2018} recently proposed a new rule splitting method. The idea is to split into two, a rule which makes too many mistakes or which is too large. However, it takes time after the splitting for the rules to re-adjust to the local distribution of points, introducing inertia in the learning. On the contrary, \cite{Song2018ASF} proposed a method to accelerate the learning of new rules after a drift based on the generation of data from new drifted concept using a GAN \cite{NIPS2014_5423}.  
\\
Similarly, we propose a complementary method to prevent instability of the rules while improving reactivity. To do so, we introduce a new original architecture of EFS, called \emph{ParaFIS}. In ParaFIS, a generalized evolving fuzzy system (principal system) is learned synchronously with \emph{an anticipation module}. For each rule of the principal system, two rules are anticipated in the anticipation module. Thus, the anticipation module enforces the system to locally adapt the distribution of points with two rules rather than just one, in order to anticipate a concept drift before it occurs.
The paper is organized as follows. Section \ref{sec:GEFS} introduces the generalized EFS and its learning model with a discussion on its drawbacks. Section \ref{sec:anticipation} introduces our contribution, the \emph{ParaFIS} evolving system. 
The experiments showing that the anticipation improves the plasticity of the system while keeping stability, are detailed in section \ref{sec:exp}. In this section, we propose to measure performance of the system on artificial brutal drifts with fitted parameters of a handcraft model. This evaluation protocol allows to quantify the time of reactivity of the system and its stability in the steady-state. 

\section{Generalized Evolving Fuzzy Systems}
\label{sec:GEFS}
In this paper, we focus on the generalized evolving fuzzy system that uses a generalized version of Takagy-Sugeno fuzzy systems, already used \cite{Almaksour2011,lemos2011,Lughofer2018}. 

\subsection{Model Architecture}
A Takagy-Sugeno (TS) fuzzy system is a set of fuzzy inference rules $R=\{r_i, 0\leq i \leq N\}$ with an antecedent part (also called premise), and a consequent part. Each rule's antecedent is defined with a prototype that is set by a cluster with a center $\mu_i$. The structure of a rule $r_i$, is as follows:
\begin{equation}
\textbf{IF } \text{ \textbf{x} is close to $\mu_i$ } \textbf{   THEN   } \text{ $y_i^1=l_i^1(\textbf{x})$ .. $y_i^c=l_i^c(\textbf{x})$  } 
\label{eq:structRule}
\end{equation}
With $l_i^j$ a polynomial function for $r_i$ of class $j$; $c$ the number of class and $N$ the number of rules. The degree of the polynomial function is set to $1$ with $\pi_{ik}^j$ the polynomial coefficients (see Eq. (\ref{eq:polfunction})). 
\begin{equation}
y_i^j=l^j_i(\textbf{x})= \pi_{i0}^j + \pi_{i1}^j x_1  + \pi_{i2}^j x_2 + .. + \pi_{in}^j x_n = \textbf{$\Pi_i^j$} \textbf{x}
\label{eq:polfunction}
\end{equation}
The membership of \textbf{x} to a rule $r_i$, denoted $\beta_i(\textbf{x})$, is given by a normalized Radial Basis Function $K$ (RBF), of the distance from $x$ to $\mu_i$ (see Eq.(\ref{eq:RBF})). The RBF is often a multivariate Gaussian or Cauchy function \cite{Almaksour2011,lemos2011}. 
\begin{equation}
\beta_i(\textbf{x})=K(||\textbf{x}-\mathbf{\mu_1} ||^2)
\label{eq:RBF}
\end{equation}
In the generalized version of TS, the Mahalanobis distance is used to get rotated hyper-ellipsoid clusters as follows:
\begin{equation}
    ||\textbf{x}-\mathbf{\mu_i}||^2=(\textbf{x}-\textbf{$\mu$}_i)A^{-1}(\textbf{x}-\textbf{$\mu$}_i)^T
    \label{eq:mahalanobis}
\end{equation}
With $A$ the covariance matrix.
Finally, the predicted class for \textbf{x} is given by Eq. (\ref{eq:predictedClass1}),(\ref{eq:predictedClass2}).
\begin{align}
class(\textbf{x})&=y=argmax_j \ y^j(\textbf{x})
\label{eq:predictedClass1} \\
\text{Where } y^j(\textbf{x})&=\sum_{i=1}^{N}\beta_i(\textbf{x})y_i^j
\label{eq:predictedClass2} 
\end{align}

\subsection{Rule's adaptation}
Each new incoming data $\textbf{x}_t$ is used to adapt the model parameters. In the premise part, only the most activated rule adapts its center and covariance matrix according to Eq. (\ref{eq:centerUpdate}),(\ref{eq:covUpdate}) where $t$ is the number of samples of the rule. 
\begin{align}
\textbf{$\mu_t$}&= \frac{t-1}{t} \mu_{t-1} + \frac{1}{t} \textbf{x}_t
\label{eq:centerUpdate}
\\
\textbf{A$_t$}&= \frac{t-1}{t} \textbf{A}_{t-1} + \frac{1}{t} (\textbf{x}_t-\mu_t)(\textbf{x}_t-\mu_t)^T
\label{eq:covUpdate}
\end{align}
The forgetting capacity is put in the equation, by setting $t=min(k,tmax)$ (see \cite{Manuel2013}) with $tmax$ a threshold that defined the forgetting capacity, and $k$ the number of samples that activated the most the rules. $tmax$ is often written with a forgetting factor $\alpha=\frac{tmax-1}{tmax} \in [0,1]$ where $\alpha=1$ when there is no forgetting capacity.
\\
The consequent part is learned using a Weighted Recursive Least Square method (WRLS).
The membership functions $\beta$ is assumed to be almost constant to converge to the optimal solution. To reduce computation time, the local learning of the consequent part is often preferred. And, the rules are assumed to be independent to apply RLS on each one. 
The conclusion matrix $\Pi_{i(t)} = [\Pi_{i(t)}^1,..,\Pi_{i(t)}^c]$ of the rule $r_i$ at time $t$ (\textit{i.e.} after $t$ data points) is recursively computed according to:
\begin{align}
\Pi_{i(t)} &=\Pi_{i(t-1)} + C_{i(t)} \beta_i(\textbf{x})C_i \textbf{x} (Y_t-\textbf{x}\Pi_{i(t-1)})  
\label{eq:consUpdate}
\\
\text{Where } C_{i(t)} &= C_{i(t-1)}-\frac{\beta_i(\textbf{x}) C_{i(t-1)} \textbf{x} \textbf{x}^T C_i}{1+\beta_i(\textbf{x})\textbf{x}^T C_i \textbf{x} }
\label{eq:corrUpdate}
\end{align}
With $C_i$ a correlation matrix initialized by $C_{i(t=0)}=\Omega Id$ where $Id$ is the identity matrix and $\Omega$ a constant often fixed to $100$ (see \cite{Almaksour2011},\cite{angelov2004}).

\subsection{Rule creation condition}
The adaptation of the parameters is not relevant to handle brutal drifts, as when a class must be represented by several clusters or when the target concept shift in the feature space. Dealing with brutal drifts requires the adaptation of the fuzzy system structure, like rule addition.
In a context of online learning from scratch, all classes start with one prototype (\textit{i.e.} one rule). The structure adaptation relies on specific conditions observed on data, via the existing rules. Several criteria are defined to detect brutal drifts.
Most EFS uses a distance-based criteria \cite{Lughofer2018},\cite{Almaksour2011} that compares a certain threshold $T(P)$ (depending on parameters $P$) with the distance between the prototype of the closest rule (its center $\mu$) and the new incoming points $\textbf{x}_t$. If $T(P)<dist(\mathbf{\mu}-\textbf{x}_{\textbf{t}})$, then the rule creation criteria is met, a new rule is created over the last incoming point according to Eq. (\ref{eq:initRuleGEFS}). 

\begin{equation}
\begin{split}
    &\mu_{new}=\textbf{\textbf{x}$_{\textbf{t}}$} \hspace{2cm} \Pi_{new}=\textbf{0} \\
    &cov_{kl}= \frac{1}{100} \delta_{kl} \hspace{1cm} \forall k,l \in [1,n]
\end{split}
\label{eq:initRuleGEFS}
\end{equation}

\subsection{Discussion on problems in the generalized EFS}

\begin{figure}
\centering
\includegraphics[width=\linewidth]{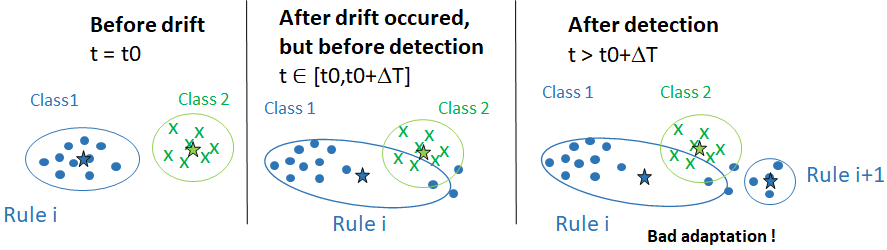}
\caption{Problems met in generalized EFS when brutal drift occurs. The antecedent structure of each rule is represented by the ellipse (the covariance matrix) and the star (the center)}
\label{fig:probGEFS}
\end{figure}

\begin{figure}
    \centering
    \includegraphics[width=0.7\linewidth]{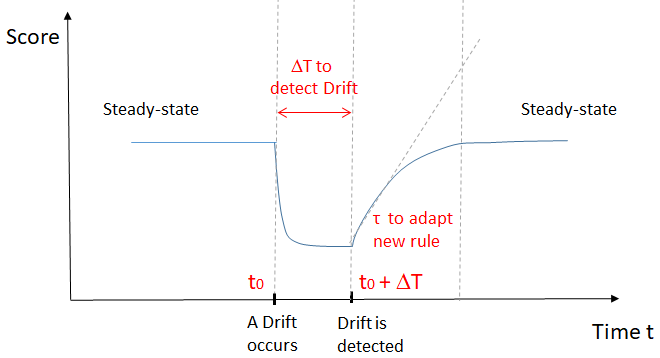}
    \caption{Impact of a drift on the score in time}
    \label{fig:driftModel}
\end{figure}

Two scales of adaptation co-exist in EFS, the adaptation of rule parameters and the structure adaptation. Smooth drifts are tackled by introducing forgetting capacity in the parameter adaptation whereas brutal drifts are tackled by the creation of new rules. 
However, the system is degraded by the parameter adaptation when a brutal drift occurs. 
Indeed, all rule creation conditions lead to a trade-off between speed of detection and sensitivity to noise. But, for all, it exists a time $\Delta T>0$ between the true occurrence of the drift, and the detection time. As shown in Figure \ref{fig:probGEFS} and \ref{fig:driftModel}, during this $\Delta T$ time, one (ore more) rule tries to adapt to the drift and changes its parameters making it less fit to its previous concept. But, when the rule creation is triggered, this previous adaptation is not cancelled making the old rule perhaps unstable. Moreover, the new rule is created over one single point (the one that triggers the rule creation) although all points during $\Delta T$ could have been used to initialize the new rule. This results in a longer time $\tau$ for the new rule adaptation to the new concept.


\section{Contribution: ParaFIS system}
\label{sec:anticipation}

In order to attend stability and plasticity, we present in this section the ParaFIS system. 

\subsection{Model's architecture}
\begin{figure}
\centering
\includegraphics[width=\linewidth]{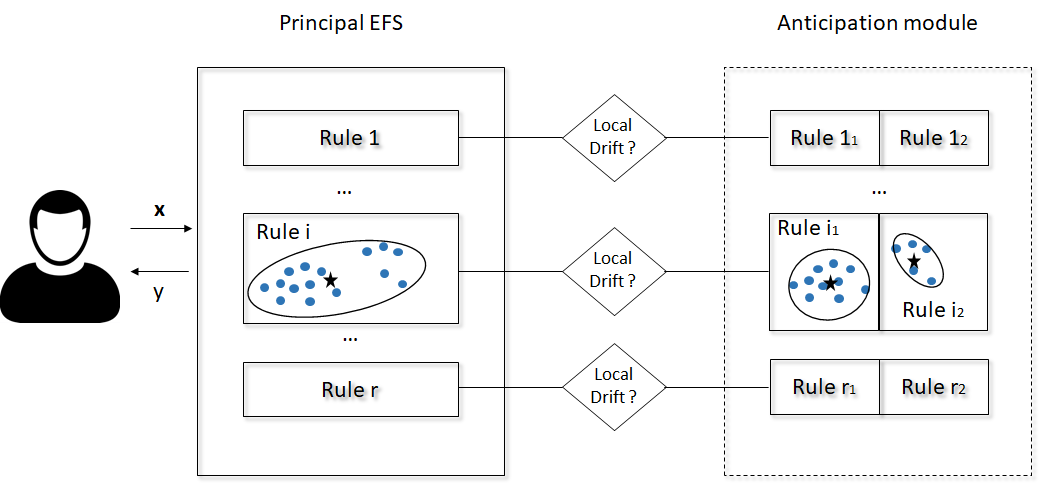}
\caption{Architecture of our proposition ParaFIS}
\label{fig:architectureParaFIS}
\end{figure}
As shown in Figure \ref{fig:architectureParaFIS}, the proposed model is based on the generalized EFS similar to this described section \ref{sec:GEFS}. This part is dedicated to smooth drifts to keep stability (no structural change). And, for each rule $r_i$, a module of anticipation is added to deal with brutal drifts. This anticipation module is composed of two sub-rules $r_{i_1}$ and $r_{i_2}$ that have an antecedent part (a center, a covariance matrix, a Cauchy membership function) and a consequent part with $c$ hyperplanes ($c$ the number of classes). The system classifies the data at any time independently of the anticipation module as it is done in the generalized EFS. The sub-rules are just used in the learning phase where different forgetting factors are applied to adapt differently the distribution of points in time. Thus, the system gets information at different scales of time.

\subsection{Model's learning}

Each new point coming into the system will be used to learn both the principal system and the anticipation module. As in the generalized EFS, the principal system will adapt the antecedent part of the most activated rule $r_{m}$ by updating its center and its covariance matrix using Eq. (\ref{eq:centerUpdate}),(\ref{eq:covUpdate}) with a factor $\alpha=1$ (no forgetting capacity). Then, the sub-rules $r_{m_1}$ and $r_{m_2}$ are updated using the same equations with $\alpha1,\alpha2 \leq 1$. The two sub-rules have two different forgetting factors leading to two temporal scales of learning. $r_{m_1}$ has a low forgetting factor and is learned on the most recent data whereas $r_{m_2}$ has a high forgetting factor and is learned on a long history. In this way, $r_{m_1}$ quickly reacts to a change in the distribution of points whereas $r_{m_2}$ preserves the old concept with a slow adaptation.
\\
The consequent part in the principal system is learned as usual (Eq. (\ref{eq:consUpdate}),(\ref{eq:corrUpdate})). In the anticipation module, $r_{i_1}$, $r_{i_2}$ have the same consequent part as $r_i$ (the $c$ hyperplanes $y_i^j$ with $j\in [1,c]$).

\subsection{Detection of brutal concept drifts}

\begin{figure}
    \centering
    \includegraphics[width=3cm]{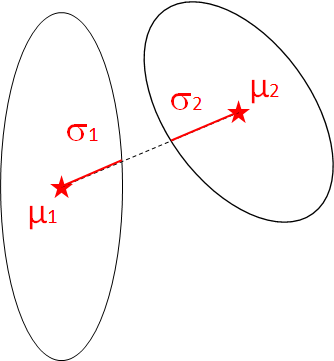}
    \caption{Separability criteria using covariance of both clusters}
    \label{fig:separabilityCriteria}
\end{figure}

Contrary to current rule creation criterion which use no information of neighborhood \cite{Lughofer2018},\cite{Almaksour2011}, we propose here to integrate a brutal drift detector based on a clustering separability criteria. The idea is to assume that if the two sub-rules in the anticipation module are enough separated, then a brutal drift occurred. Then, $r_{m_2}$ learned over the large history matches the old concept and $r_{m_1}$ learned over the few last points matches the new drifted concept.
Eq. (\ref{eq:Condition1}) presents the proposed separability criteria that is based on the covariance of both clusters.
 \begin{equation}
 \textbf{Condition 1} \hspace{1.2cm}  ||\mu_i-\mu_j|| > \sigma_i + \sigma_j \hspace{1cm} 
    \label{eq:Condition1}
 \end{equation}
 Where, as depicetd in Figure \ref{fig:separabilityCriteria}, $\sigma_i$ (resp. $\sigma_j$) is the distance between $\mu_i$ (resp. $\mu_j$) and the hyper-ellipsoid's envelop of cluster $i$ (resp. $j$), along $(\mu_i,\mu_j)$ axis.
Besides, to force the rules to take into account a certain number of points $n_{min}$ (20 by default) before deciding to create a new rule, the following inertia criteria is added:
\begin{equation}
\textbf{Condition 2} \hspace{2cm}  k_i> n_{min} \hspace{2cm} 
\label{eq:Condition2}
\end{equation}

\subsection{Initialization of new rules based on the anticipation module}

If the rule creation conditions are met for the most activated rule $r_m$ from the principal system, then $r_m$ is replaced by $r_{m_1}$ and $r_{m_2}$ in the principal system. Then for each new $r_{m_i}$ of the principal system, two new sub-rules $r_{m_{i_1}}$, $r_{m_{i_2}}$ are initialized in the anticipation module, as follows: 
\begin{equation}
\begin{split}
    &\mu_{r_{m_{i_1}}} =  \mu_{r_{m_{i_2}}}=\mu_{r_{m_i}} \hspace{1cm} A_{r_{m_{i_1}}} = A_{r_{m_{i_2}}} =A_{r_{m_i}} \\
    &\Pi_{r_{m_{i_1}}}=\Pi_{r_{m_i}} \hspace{0.8cm}  \Pi_{r_{m_{i_2}}}=0 \hspace{1cm} k_{r_{m_{i_1}}}=k_{r_{m_{i_2}}}=0
\end{split}
\label{eq:initSubRules} 
\end{equation}
\\
The idea is to keep the learned information of $r_{m_i}$ in the sub-rule $r_{m_{i_1}}$, and to initialize the second $r_{m_{i_2}}$ from scratch.
All steps of the learning are summarized in Algorithm~\ref{algo}.

\begin{algorithm}
\caption{Learn ParaFIS model}
\begin{small}
\begin{algorithmic}
\REQUIRE $x_{new}$, current fuzzy rule system containing set of rules R and set of classes C
\IF{$class(x_{new}) \in C$}
    \STATE Find max activated rule $r_m \in$ R according to (\ref{eq:RBF})
    \STATE Check the \textbf{condition 1} and \textbf{condition 2} Eq. (\ref{eq:Condition1}-\ref{eq:Condition2}) with $r_m$
    \IF{Condition 1 \& Condition 2 == FALSE } 
        \STATE According to Eq.(\ref{eq:centerUpdate})-(\ref{eq:covUpdate}), update the antecedent part of $r_m$ with $\alpha =1$, and $r_{m_1}$, $r_{m_2}$ with $\alpha_1 \neq \alpha_2$
        \STATE According to Eq. (\ref{eq:consUpdate})-(\ref{eq:corrUpdate}), update the consequent part of all rules $r_i$, $r_{i_{1}}$ and $r_{i_{2}}$
    \ELSE 
        \STATE Replace $r_m$ by $r_{m_1}$,$r_{m_2}$ in the principal system
        \STATE Initialize 4 new sub-rules in the anticipation module Eq. (\ref{eq:initSubRules})
    \ENDIF
\ELSE
    \STATE Create a new rule Eq. (\ref{eq:initRuleGEFS}) 
    \STATE Initialized two sub-rules in the anticipation module with (\ref{eq:initRuleGEFS})

\ENDIF
\end{algorithmic}
\end{small}
\label{algo}
\end{algorithm}

\section{Experimental Validation}
\label{sec:exp}

\subsection{Evaluation protocol}

\subsubsection{Prequential test with artificial brutal drift}

\begin{figure}
\centering
\includegraphics[width=7cm]{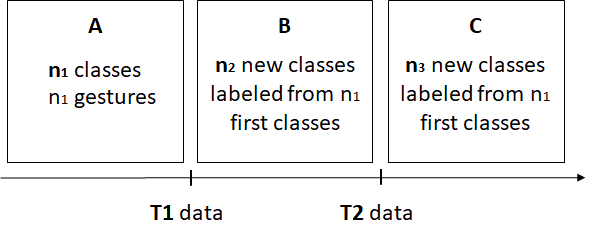}
\caption{Protocol $P$ - Generation of data stream with brutal drifts}
\label{fig:protocol}
\end{figure}

To evaluate the performance of an online classifier, it is current to use the prequential test \cite{Gama2013}. In this test, data are given one by one to the system. The system first tests the new data to get a score (1 if the class is well classified, 0 otherwise) and then learns on it. In this way, all the data are used to test the system and then, to train it while maintaining independence between each phase. Scores are then averaged over a certain window (of size $n$) to get a smooth curve of the performance over time. This test simulates a real usecase as for many online application where the system must adapt to the behavior of a user along time. In the following experiments, $n=5$ (except to plot the figures $n=100$ to smooth the score).
\\
There is no existing dataset with annotation of the nature of the drift or with the occurrence time of the drift making the evaluation and comparison of online classifiers complex. To make the task easier, the paper proposes to generate artificial brutal drifts in real data at a chosen time. 
To do so, each dataset is split into three sub-datasets $A$, $B$, $C$ to create a specific data stream. As illustrated in Figure \ref{fig:protocol}, $A$ is composed of the first $T1$ data belonging to $n_1$ classes. $B$ (resp. $C$) is composed of the data in the stream between $T1$ and $T2$ (resp. between $T2$ and $T3$), this data belongs to $n_2\leq n_1$ (resp. $n_3\leq n_1$) different classes and are relabeled with the $n_1$ labels to produce the brutal drift. 
In this ways, $n_2$ brutal drifts are done at the time $T1$ and $n_3$ others at time $T2$. Thereafter, we call this approach the \textbf{protocol $P$}. 
In the context of a command-gesture system, the protocol is equivalent to a user changing the gesture of a command while keeping the possibility to make the old gesture.

\subsubsection{Benchmark dataset}

Regarding the context of command-gestures system, three datasets from the handwritten pattern recognition community have been chosen to assess the performance of the system: PenDigits \cite{PenDigits}, Letters \cite{Letters} and LaViola \cite{Laviola}. They are all available in the UCI machine learning repository \cite{UCI}. All three datasets contain different features extracted from the handwritten patterns (digits, letters or symbols handwritten by different writers). Description of each dataset is given in Table \ref{benchmark}.
\begin{table}
\centering
\begin{tabular}{cccccc}
  \hline
  \textbf{Dataset} & \textbf{Classes} & \textbf{Features} & \textbf{Samples} & \textbf{Scriptwriters}  \\
  \hline
  \textbf{Letters}  &26&16&20000& 20  \\
  \textbf{LaViola } &48&50&16891&34\\
 \textbf{PenDigits } &10&16&10992&44\\
  \hline
\end{tabular}
\caption{Information on the datasets used in the experiments}
\label{benchmark}
\end{table}
These datasets are built in a static context meaning that there is no order between data. Thus, the prequential test with artificial brutal drifts can be done several times ($m$) by shuffling the dataset. All results given thereafter are averaged over $m$ prequential tests with $m=100$. The experimental parameters are given in Table \ref{table:expParam}. The $n1,n2,n3$ classes of each dataset follow the order of occurrence in the data file from UCI.

\begin{table}
\centering
\begin{tabular}{ccccccc}
  & \textbf{T1} & \textbf{T2-T1}  & \textbf{T3-T2} & \textbf{n1} & \textbf{n2} & \textbf{n3}  \\
  \hline
  \textbf{Letters} & 2000& 4000 & 4000 & 10 & 10 & 6  \\
   \textbf{PenDigits} & 2000& 3000 & 3000 & 4 & 3 & 3  \\
   \textbf{Laviola} & 2000& 3000 & 3000 & 10 & 10 & 10  \\
\end{tabular}
\caption{Experimental parameters}
\label{table:expParam}
\end{table}

\subsubsection{Characterization of the plasticity and stability}

\begin{figure}
\centering
\includegraphics[width=0.6\linewidth]{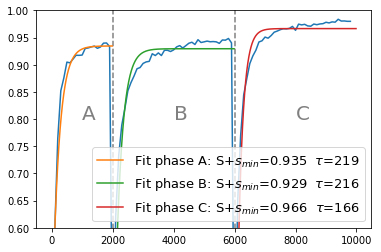}
\caption{Fit of the three phases score using Eq. (\ref{eq:fit})}
\label{Fit}
\end{figure}

In order to measure the plasticity and stability of the system, we propose to fit the prequential score with an handcraft model given by Eq. (\ref{eq:fit}).
\begin{equation}
    y(t)=S(1-e^{-\frac{t}{\tau}})+s_{min}
    \label{eq:fit}
\end{equation}
Where we define a characteristic time \textbf{$\tau$}, which represents the reactivity time (or the plasticity). In particular after a time $t=\tau$, 63\% of the score have been reached until the steady state. At the steady state, the score is given by \textbf{$S+s_{min}$}. A least square method is used to determine the parameters which best fit the curve with the model. The example of fits on the Letters dataset (experimented with the protocol P) given in Figure~\ref{Fit} shows that this model fits well the prequential score.

\subsection{Evaluation of the anticipation}

\subsubsection{Importance of covariance matrix initialization}
No study has yet tackled the issue of optimizing the initialization parameters of the antecedent part.
Yet, the initialization of the prototype (the antecedent part) greatly influence the score. To show it, we propose to compare three methods currently used in incremental fuzzy system to initialize the covariance of a new rule: Method I1 (Eq. (\ref{eq:initCov1}), \cite{Manuel2013}), Method I2 (Eq. (\ref{eq:initCov2}), \cite{Lughofer2018},\cite{lemos2011}), Method I3 (Eq. (\ref{eq:initCov1})). The center of prototypes is initialized just on the last point for all methods.\\

$\forall k,l \in [1,n] \text{, } \forall j \in [1,N]$
\begin{align}
    \textbf{I1:} \hspace{1cm} cov_{kl}&= min_j(cov^j_{kl}) \delta_{kl}  
    \label{eq:initCov1} \\
    \textbf{I2:} \hspace{1cm} cov_{kl}&= \frac{1}{100} * \delta_{kl} 
    \label{eq:initCov2} \\
    \textbf{I3:} \hspace{1cm}cov_{kl}&= \frac{mean_j(cov^j_{kl})}{10}
    \label{eq:initCov3} 
\end{align}
To compare them, we use the generalized evolving fuzzy system as described section \ref{sec:GEFS} with \textbf{Condition 1} and \textbf{Condition 2} to create rules (section \ref{sec:anticipation}). Results are obtained from the three datasets and a fit using Eq. (\ref{eq:fit}) is done.
Results are displayed in Figure \ref{fig:prequentialScore} column A, and the fitted parameters are shown in Table \ref{table:fittedParametersRandom}. Results show that the best choice of initialization depends on the dataset.
However it is clear that the initialization of the rules widely impacts the results. For instance, there is $5.6\%$ of difference for the $S+s_{min}$ parameters between methods $I3$ and $I2$ for letters dataset at the phase B. This highlights the importance of well initializing the covariance matrix of the new rule. This can be explained by the fact the covariance matrix plays a great role in the learning. Indeed, only the antecedent part of the most activated rule is learned on the new incoming point and the activation, computed with a Mahalanobis distance, crucially depends on the shape of the covariance matrix. If it is too small, few data from the new concept will activate the new rule. If it is too big, data from other concepts may also activate the new rule.

\begin{table}
\begin{small}
\centering
\begin{tabular}{p{1.4cm}p{0.1cm}|p{0.85cm}c|p{0.85cm}c|p{0.85cm}c}
  & & \multicolumn{2}{c|}{\textbf{A}}  & \multicolumn{2}{c|}{\textbf{B}} & \multicolumn{2}{c}{\textbf{C}}  \\
  & & $S +~s_{min}$ & $\tau$ & $S +~s_{min}$ & $\tau$ & $S +~s_{min}$ & $\tau$ \\
  \hline
  \textbf{Letters} &I1& \textbf{93.4} & 218 & 87.9 & 443 & 94.0 & 377 \\
   &I2& \textbf{93.4} & 215 & 83.3 & 556 & 89.8 & 697 \\
   &I3& 92.6 & \textbf{157} & \textbf{89.9} & \textbf{243} & \textbf{94.1} & \textbf{276} \\
   \hline
  \textbf{PenDigits} &I1& \textbf{98.9} & 80 & \textbf{99.3} & \textbf{29} & \textbf{98.1} & \textbf{233} \\
   &I2& 98.9 & 80 & 98.7 & 59 & 96.9 & 415 \\
    &I3& 98.9 & 77 & 98.7 & 36 & 97.4 & 235 \\
\hline
    \textbf{Laviola} &I1& \textbf{98.7} & \textbf{66} & 96.7 &88 & 96.5 & 131 \\
                             &I2& 98.7 & 68 & 96.7 &90 & 96.3 & 135 \\
                             &I3& 98.2 & 67 & \textbf{97.1} &\textbf{66} & \textbf{97.2} & \textbf{92} \\
\end{tabular}
\end{small}
\caption{Fitted parameters of the handcraft model (\ref{eq:fit})}
\label{table:fittedParametersRandom}
\end{table}

\subsubsection{Comparison with our proposition}
Now, the impact of the anticipation will be evaluated.
The antecedent part initialization is compared when using the best method among \{I1,I2,I3\} for each dataset, and using the anticipation module Eq.(\ref{eq:initSubRules}). 
However the drift detector can depend on the parameters of the system as it is in ours, Eq. (\ref{eq:Condition1}) (this depends on the covariance matrices). Thus different initialization of prototype change the times of detection. To avoid this bias, the drift detector is first used on the system with anticipation and the times $t$, at which the drifts are detected, are saved. Then, rather than using the detector, we use the saved files with all times $t$ to create rules. In such ways, the system can be compared fairly without bias of the detection.
\\
Two sets of parameters are used for the anticipation, $ Para_1 =(\alpha_1=1,\alpha_2=0.9)$ and $Para_2=(\alpha_1=1,\alpha_2=0.95)$ with $\alpha_1$ (resp. $\alpha_2$) the forgetting factor of the sub-rules $i1$ (resp. $i2$) of the secondary system for $i \in [1,r]$.
\\
Moreover, a comparison is done with our own implementation of an EFS similar to Gen-Smart EFS \cite{Lughofer2018} called GEFS*. In this last one, the rule creation criteria (given by Eq. (\ref{eq:RuleCreationCriteriaGEFS})-(\ref{eq:RuleCreationCriteriaGEFS2})) is the same than Gen-smart EFS \cite{Lughofer2018} with the same rule initialization method (I2). However, there are no rule merging or rule splitting methods as there are in Gen-Smart EFS.
\begin{equation}
    (\textbf{x}-\textbf{$\mu$}_i)A^{-1}(\textbf{x}-\textbf{$\mu$}_i)^T > r_i^2
    \label{eq:RuleCreationCriteriaGEFS}
\end{equation}
\begin{equation}
    r_i=\kappa p^{\frac{1}{\sqrt2}} \frac{1.0}{(1-\frac{1}{k_i+1})^m}
    \label{eq:RuleCreationCriteriaGEFS2}
\end{equation}

\begin{figure}
    \centering
    \includegraphics[width=\linewidth]{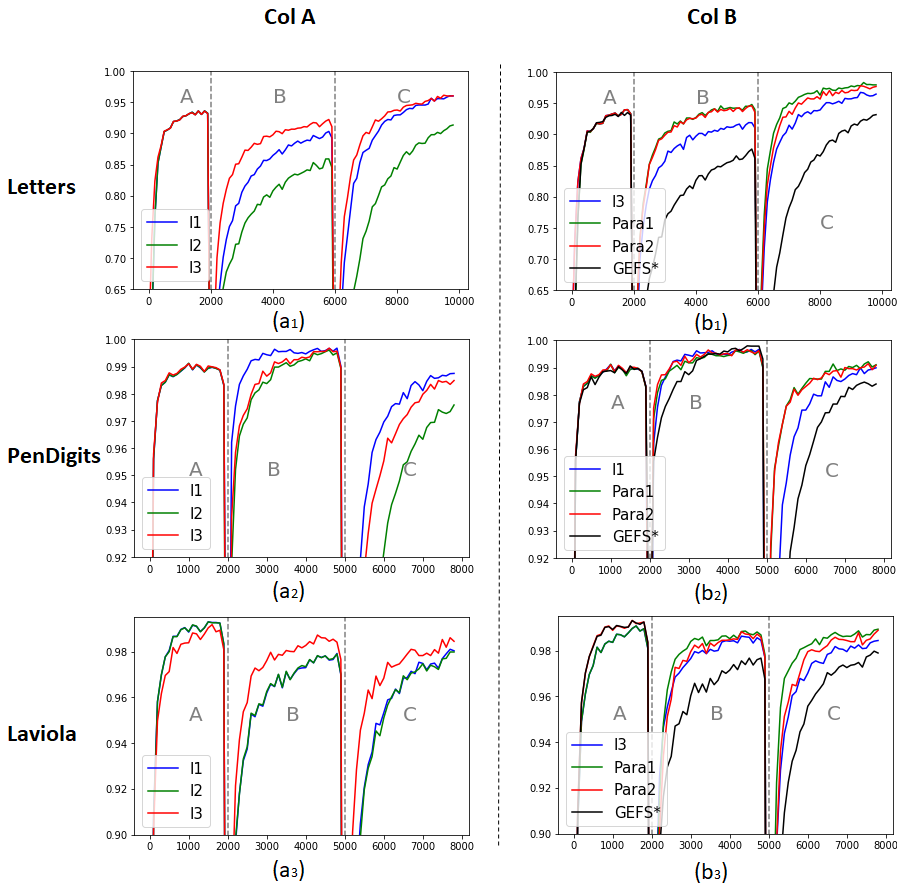}
    \caption{Prequential score (y-axis) with respect to data (x-axis) for the three datasets. The first (resp. the second) column refers to the first (resp. second) experiment}
    \label{fig:prequentialScore}
\end{figure}

Results are displayed in Figure \ref{fig:prequentialScore} column B, and Table \ref{table:meanScore}. The fitted parameters of each configuration are averaged over the three phases. The mean accuracy score is also given to compare global performance of each configuration.
The results show that anticipation brings a gain in the mean score, not only on the reactivity with a lower time $\tau$, but also on the score reached in the steady state. This results in a better mean accuracy score. It means that the anticipation effectively accelerates the learning of the new concept when new rule are created, but also stabilizes the covariance matrix to better match the target concept at end. Moreover, our global system ParaFIS with detector+anticipation outperforms, in term of reactivity and stability, an equivalent system GEFS* with detector+initialization from the state of the art \cite{Lughofer2018}.

\begin{table}
\centering
\begin{tabular}{cc|cc|c}
  & & $<S+s_{min}>$ & $<\tau>$ & $<Acc>$ \\
  \hline
  \textbf{Letters } &$Para_1$& \textbf{94.3} & 200 &\textbf{90.3} \\
   &$Para_2$& 93.8 & \textbf{188} & 90.1 \\
   &I3& 92.2 & 214 &   88.1\\
    &GEFS*& 89.9 & 471 & 80.7 \\
    & \tiny{$\kappa=1.6$ , $m=4$} & &  \\
   \hline
  \textbf{PenDigits } &$Para_1$& 98.8 & \textbf{56} &   97.9  \\
   &$Para_2$& \textbf{98.9} & \textbf{56} & \textbf{98.0}  \\
    &I1& 98.8 & 103 & 97.2  \\
    &GEFS*& 98.7 & 156 & 96.3 \\
    & \tiny{$\kappa=2.6$ , $m=4$} & & \\

\hline
    \textbf{Laviola } &$Para_1$& \textbf{98.2} & \textbf{74} & \textbf{96.2} \\
     &$Para_2$& \textbf{98.2} & 84 &  95.9\\
    &I3& 97.7 & 76 & 95.7  \\
    & GEFS*& 97.1 & 158 & 94.4 \\
    & \tiny{$\kappa=2.5$ , $m=4$} & &  \\
\end{tabular}
\caption{Mean score over the three phases A,B,C}
\label{table:meanScore}
\end{table}

\section{Conclusion \& Outlooks}
This paper has introduced a new design of EFS which integrates an anticipation module to deal with brutal drifts.
Anticipation opens up a new interesting way to tackle non stationary environment problems. It has allowed to detect brutal drifts and adapt new rules to the drifted concept with a gain in reactivity and stability. The new design opens up to new outlooks with the two sub-rules that map the rule creation problem into a 2-cluster clustering problem. It may use other online clustering validity criteria to detect brutal drifts or add new anticipation modules to tackle other natures of drifts such as gradual drifts (where data switch from one target concept to another one several times).

\bibliographystyle{unsrt} 
\bibliography{biblio}

\begin{thebibliography}{10}

\bibitem{Bouillon2017Nov}
M.~Bouillon and E.~Anquetil.
\newblock {Online active supervision of an evolving classifier for
  customized-gesture-command learning}.
\newblock {\em Neurocomputing}, 262:77--89, Nov 2017.

\bibitem{Lughofer2018}
E.~Lughofer, M.~Pratama, and I.~Skrjanc.
\newblock Incremental rule splitting in generalized evolving fuzzy systems for
  autonomous drift compensation.
\newblock {\em IEEE Transactions on Fuzzy Systems}, 26(4):1854--1865, Aug 2018.

\bibitem{angelov2011}
P.~Angelov and R.~Yager.
\newblock {Simplified fuzzy rule-based systems using non-parametric antecedents
  and relative data density}.
\newblock {\em 2011 IEEE Workshop on Evolving and Adaptive Intelligent Systems
  (EAIS)}, pages 62--69, 2011.

\bibitem{Liu2017Jul}
A.~Liu, G.~Zhang, and J.~Lu.
\newblock {Fuzzy time windowing for gradual concept drift adaptation}.
\newblock {\em 2017 IEEE International Conference on Fuzzy Systems
  (FUZZ-IEEE)}, pages 1--6, Jul 2017.

\bibitem{Manuel2013}
M.~Bouillon, E.~Anquetil, and A.~A. Almaksour.
\newblock Decremental learning of evolving fuzzy inference systems using a
  sliding window.
\newblock In {\em Proceedings of the 2012 11th International Conference on
  Machine Learning and Applications - Volume 01}, ICMLA '12, pages 598--601,
  Washington, DC, USA, 2012. IEEE Computer Society.

\bibitem{Song2018ASF}
Y.~Song, G.~Zhang, H.~Lu, and J.~Lu.
\newblock A self-adaptive fuzzy network for prediction in non-stationary
  environments.
\newblock {\em 2018 IEEE International Conference on Fuzzy Systems
  (FUZZ-IEEE)}, pages 1--8, 2018.

\bibitem{NIPS2014_5423}
I.~Goodfellow, J.~Pouget-Abadie, M.~Mirza, B.~Xu, D.~Warde-Farley, S.~Ozair,
  A.~Courville, and Y.~Bengio.
\newblock Generative adversarial nets.
\newblock In {\em Advances in Neural Information Processing Systems 27}, pages
  2672--2680. Curran Associates, Inc., 2014.

\bibitem{Almaksour2011}
A.~A. Almaksour and E.~Anquetil.
\newblock Improving premise structure in evolving takagi--sugeno neuro-fuzzy
  classifiers.
\newblock {\em Evolving Systems}, 2(1):25--33, Mar 2011.

\bibitem{lemos2011}
A.~Lemos, W.~Caminhas, and F.~Gomide.
\newblock Multivariable gaussian evolving fuzzy modeling system.
\newblock {\em IEEE Transactions on Fuzzy Systems}, 19(1):91--104, Feb 2011.

\bibitem{angelov2004}
P.~P. Angelov and D.~P. Filev.
\newblock An approach to online identification of takagi-sugeno fuzzy models.
\newblock {\em IEEE Transactions on Systems, Man, and Cybernetics, Part B
  (Cybernetics)}, 34(1):484--498, Feb 2004.

\bibitem{Gama2013}
J.~Gama, R.~Sebasti{\~a}o, and P.~P. Rodrigues.
\newblock On evaluating stream learning algorithms.
\newblock {\em Machine Learning}, 90(3):317--346, Mar 2013.

\bibitem{PenDigits}
F.~Alimoglu.
\newblock {\em Combining Multiple Classifiers for Pen-Based Handwritten Digit
  Recognition}.
\newblock PhD thesis, Bogazici Univeristy, 1996.

\bibitem{Letters}
P.~W. Frey and D.~J. Slate.
\newblock {Letter recognition using Holland-style adaptive classifiers}.
\newblock {\em Mach. Learn.}, 6(2):161--182, Mar 1991.

\bibitem{Laviola}
J.~J. LaViola and R.~C. Zeleznik.
\newblock A practical approach for writer-dependent symbol recognition using a
  writer-independent symbol recognizers, November 2007.

\bibitem{UCI}
D.~Dheeru and E.~Karra~Taniskidou.
\newblock {UCI} machine learning repository, 2017.

\end{thebibliography}

\end{document}